# Evaluating the Performance of Deep Learning Models in Whole-body Dynamic 3D Posture Prediction During Load-reaching Activities

Seyede Niloofar Hosseini†, Ali Mojibi†, Mahdi Mohseni, Navid Arjmand, and Alireza Taheri

***Abstract*— This study aimed to explore the application of deep neural networks for whole-body human posture prediction during dynamic load-reaching activities. Two time-series models were trained using bidirectional long short-term memory (BLSTM) and transformer architectures. The dataset consisted of 3D full-body plug-in gait dynamic coordinates from 20 normal-weight healthy male individuals each performing 204 load-reaching tasks from different load positions while adapting various lifting and handling techniques. The model inputs consisted of the 3D position of the hand-load position, lifting (stoop, full-squat and semi-squat) and handling (one- and two-handed) techniques, body weight and height, and the 3D coordinate data of the body posture from the first 25% of the task duration. These inputs were used by the models to predict body coordinates during the remaining 75% of the task period. Moreover, a novel method was proposed to improve the accuracy of the previous and present posture prediction networks by enforcing constant body segment lengths through the optimization of a new cost function. The results indicated that the new cost function decreased the prediction error of the models by approximately 8% and 21% for the arm and leg models, respectively. We indicated that utilizing the transformer architecture, with a root-mean-square-error of 41.4 mm, exhibited approximately 58% more accurate long-term performance than the BLSTM-based model. This study merits the use of neural networks that capture time series dependencies in 3D motion frames, providing a unique approach for understanding and predict motion dynamics during manual material handling activities.***

This work was supported by grants from Sharif University of Technology, Tehran, Iran

*(Corresponding author: Alireza Taheri).*

S. N. Hosseini and A. Mojibi contributed equally.

The authors are with the Department of Mechanical Engineering, Sharif University of Technology, Tehran, Iran (email: niloofar.hosseini22@sharif.edu; ali.mojibi99@sharif.edu; m.mohseny2014@gmail.com; arjmand@sharif.edu; artaheri@sharif.edu ).

This work involved human subjects in its research. The data collection was carried out after institutional ethics committee approval (approval ID: IR.IUMS.REC.1401.740) and informed consent from subjects.

## I. INTRODUCTION

MANUAL material handling (MMH) activities have been identified as a risk factor for spinal injuries [1]. To assess the risk of injury in different daily activities, practitioners in the field of occupational biomechanics need to estimate external loads and moments in intervertebral joints. As direct measurements of these loads and moments are costly and invasive, musculoskeletal models and ergonomic-based risk assessment tools have been developed for this purpose [2], [3], [4], [5]. These biomechanical models require, as inputs, the individual's body posture data during the occupational activity under consideration. Analyzing human body movement by traditional methods such as motion capture cameras [6], [7], [8], [9], [10] or wearable sensors [11], [12], [13] is also costly, time-consuming, subject to occlusion, and therefore difficult to implement in real workstations. Developing easy-to-use yet accurate human posture prediction approaches during such occupational activities is, hence, essential.

The analysis of human movement, particularly for tasks involving MMH activities, has significantly benefited from the integration of computational techniques like computer vision (CV) and machine learning (ML) [14], [15], [16]. Several posture prediction models for MMH activities have been developed by our research group [12], [17], [18], [19], [20], [21], [22] and others [23], [24], [25]. In all these models, body motion data of several subjects are first measured in laboratory settings, and subsequently, machine learning-based prediction models are trained and validated using the collected data. The input parameters of these models are easily measurable and applicable in occupational environments. Some studies [12], [18], [19], [20], [21], [23] have used regular fully connected neural networks (FCNs) to predict posture, while others [24], [26] have employed generative models such as conditional variational autoencoders and conditional generative adversarial networks. Moreover, the performance of generative models has been compared to that of an FCN, both having similar errors in predicting body posture [24]. However, these models do not accurately capture the time-varying nature of dynamic load-reaching or load-moving activities. Moreover, almost all earlier posture prediction models [12], [17], [20], [21], [25] were developed for static tasks or assumed temporal independence for each movement frame for dynamic tasks [18], [19].

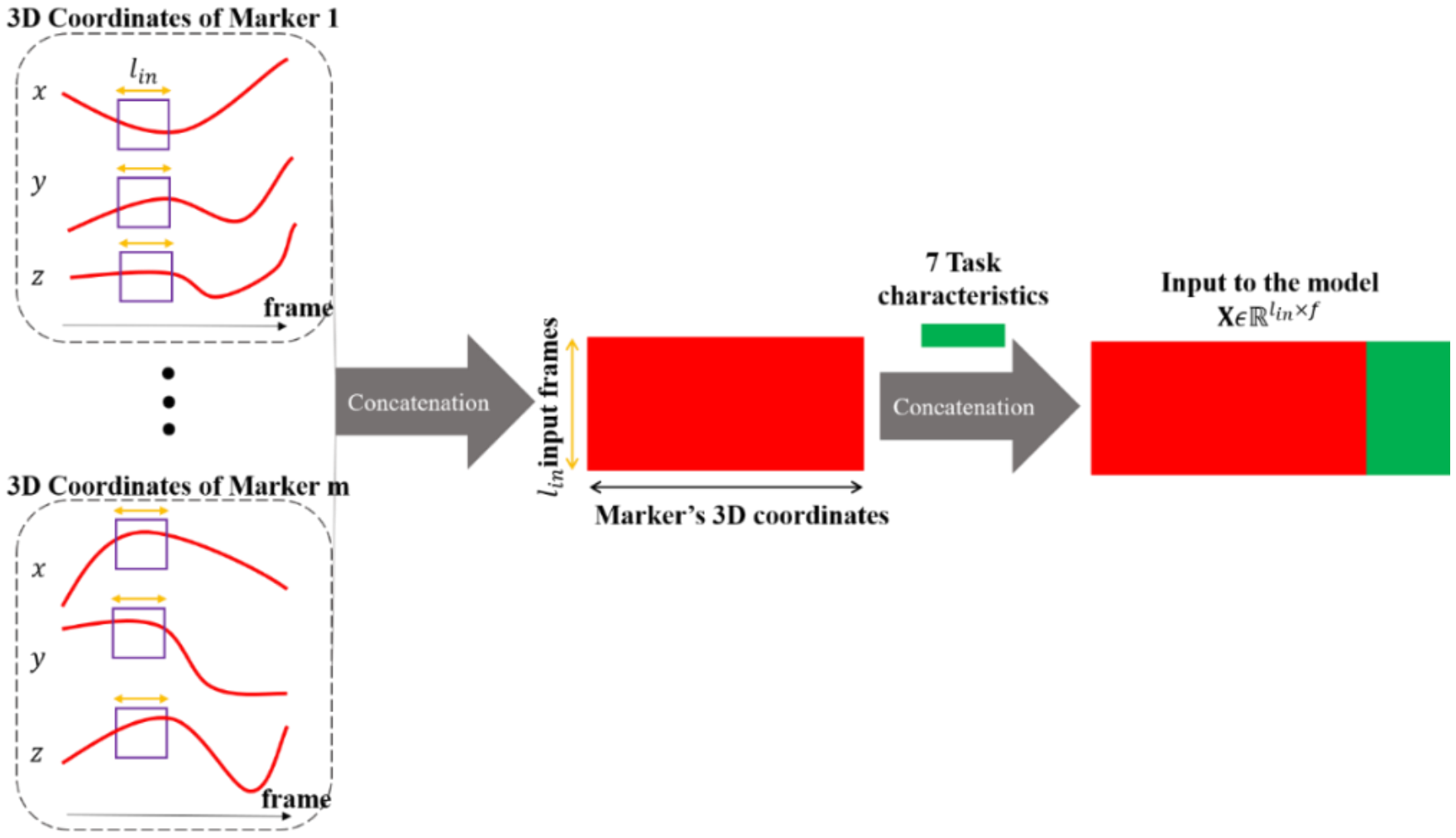

**Fig. 1.** The process of generating inputs to the neural networks from time series data: for each segment, the 3D coordinates of the markers in $l_{\text{in}}$ = 25 input frames and the task characteristics are combined to generate one sample for input to the networks.

This means that they did not consider any correlation or relationship between the movement frames as time series.

Recent advancements in deep learning, however, offer potential solutions to address these temporal limitations [14]. For instance, long short-term memory (LSTM) models have been used to predict gait trajectory [27], [28] and a comparative analysis of LSTM, FCN, convolutional neural networks (CNNs), and transformer deep learning models was conducted for predicting gait trajectories [29]. It is concluded that the FCN and transformer outperformed the other models in long-term prediction due to their stability and low error rates [29]. Therefore, transformer networks are recommended for predicting biomechanical time series [29]. Furthermore, some studies have explored enhancing LSTM models with attention mechanisms to improve the performance [30], [31], [32]. S. Li et al. [31] and Zhu et al. [30] introduced attention layers at the input stage to weight features before passing them into the LSTM for joint angle prediction. Similarly, Ding et al. [32] demonstrated that an attention-enhanced LSTM network could estimate ankle and hip acceleration prediction from inertial measurement unit (IMU) signals by leveraging complementary limb data with 9.06% and 7.64% normalized root mean square error (nRMSE). Wu et al. [33], developed a novel time series forecasting method using a Transformer architecture with self-attention mechanisms, achieving a root mean square error (RMSE) of 0.605 in influenza-like illness forecasting using time delay embeddings. These findings suggest that both deep learning architectures - particularly those utilizing recurrent structures (such as LSTM and bidirectional-LSTM (BLSTM)) and attention-based models (transformer) - are strong candidates for predicting biomechanical time series data.

Therefore, although deep learning models have been widely used to predict one-dimensional (1D) periodic biomechanical signals such as joint angles[27], [28], [29], [34], to the best of our knowledge, they have not been applied to the prediction of full-body, three-dimensional (3D), and non-periodic posture data. This presents a gap in the current literature, particularly for dynamic and complex movements. The present study aims to predict human posture during load-reaching activities as time series data using BLSTM and transformer deep neural networks to evaluate their prediction accuracy and compare their outcomes to those of FCN-based models. Moreover, kinematic constraints based on the length of body segments (i.e., forearms, arms, and shanks) are included in the loss function and their effects on network performance are investigated. It is hypothesized that these models are robust tools for predicting dynamic 3D full-body posture.

## II. Methods

### *A. Data collection*

The motion dataset collected in our previous study was used [20]. Twenty healthy right-handed male individuals (73.7±10.1 kg, 177.8±4.2 cm, 24±2 years) without any history of musculoskeletal disorders in the last six months each performed 204 load-reaching tasks (Table I). Ten Vicon motion capture cameras measured kinematics of the subjects' body movement during each load-reaching activity (i.e., reaching the load destination from the neutral upright standing posture) at a sampling frequency of 120 Hz. Thirty-nine markers were placed on each subject's body landmarks based on the full-body plug-in gait standard [35], [36] while two additional markers were placed on their T12 and S1 vertebrae [20]. These additional markers helped determine the kinematics of the spine more accurately. The measured trajectories of these markers for each load-reaching task were subsequently resampled in 101 frames to have the same length between tasks and subjects.

### *B. Data preprocessing*

3D coordinate data were filtered using a zero-lag phase 4$^{th}$ order low-pass Butterworth filter with a 10 Hz cut-off frequency. All data points were within expected biomechanical ranges and showed no anomalies that would justify removal. The data from 20 subjects were then randomly divided into the training, validation, and test sets (14, 4, and 2 subjects, respectively). The training dataset was used to update the parameters of the model, the validation dataset was used to adjust the hyperparameters of the model, and the performance of the model was reported using the test dataset. The samples for each dataset were generated using the conventional sliding window method with a length of 25 frames (i.e., 3D coordinates of 25 frame motions) and a stride of 1 frame [28].

**TABLE I**. Tasks performed by each subject at various horizontal and vertical hand locations while adapting different lifting (LT) and handling (HT) techniques [20]. LTs included upright standing (back and knees straight), stoop (knees straight and back bent), full-squat (knees bent and back straight), and semi-squat (back and knees bent). HT included one- and two-handed activities.

| Handling technique (HT) | Lifting technique (LT) | Horizontal positions of the hand(s) ($x$, $y$) from the midpoint of heels (cm) | Vertical positions of the hand(s) from the floor (cm) | Number of tasks | Total number of tasks |
|---|---|---|---|---|---|
| **One-handed** | Upright standing or stoop | (30,0), (45,0), (60,0), (30,30),(45,45), (60,60), (0,30), (0,45), (0,60), (-30,30),(-45,45), (-60,60), (-30,0), (-45,0), (-60,0) | 0, 30, 60, 90, 120, 150, 180 | 105 | 105 one-handed tasks |
| | Semi-squat | - | - | 0 | |
| | Full-squat | - | - | 0 | |
| **Two-handed** | Upright standing or stoop | (30,0), (45,0), (60,0), (30,30), (45,45), (60,60), (0,30), (0,45), (0,60) | 0, 30, 60, 90, 120, 150, 180 | 63 | 99 two-handed tasks |
| | Semi-squat | | 0, 30 | 18 | |
| | Full-squat | | | 18 | |
| **Total** | | **204 one-and two-handed unloaded tasks** | | | |

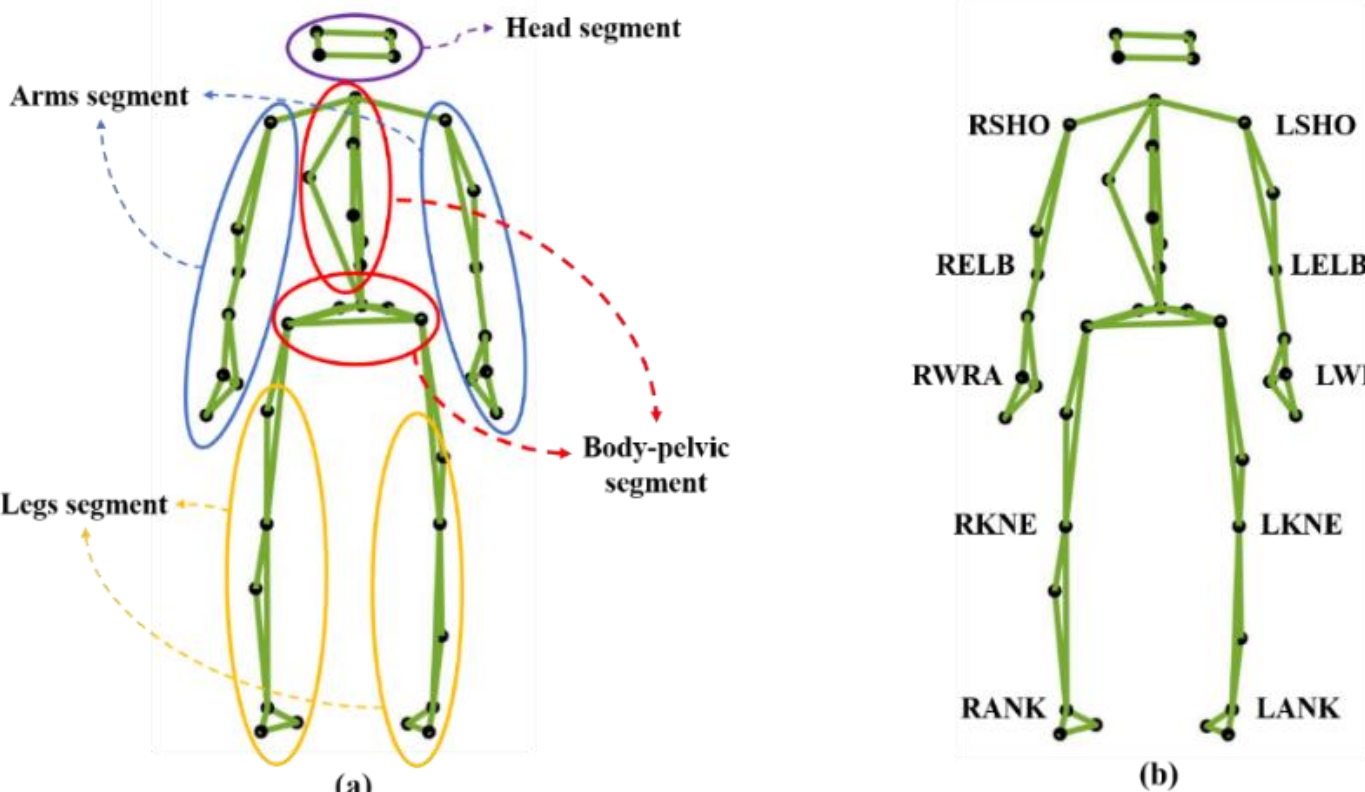

**Fig. 2.** Skeleton for the human body with attached markers: (a) four different segments used for posture prediction and (b) important markers.

A sample **X** (Fig. 1) in each set was a matrix of shape $l_{in} \times f$ ,$l_{in}$ is the number of input frames (set to 25 in this study), and $f$ is the number of features. These features included the 3D coordinates of body markers, along with seven task- and subject-related characteristics: the 3D final coordinates of the load-reaching position, the handling technique (one- or two-handed), the lifting technique (upright standing, semi-squat, full squat, or stoop), and the subject's body height and mass. The networks received integer-encoded values for the lifting and handling techniques. Specifically, 0, 1, 2, and 3 represented upright standing, stoop, semi-squat, and full-squat lifting techniques, respectively, while 1 and 2 indicated one- and two-handed handling techniques.

The label that corresponds to a sample **X** was represented by a vector $y$, with a shape of $3m{\times}1$ in which $m$ denotes the number of markers attached to the body. The vector contains the 3D coordinates of the markers in the next frame. The generated samples were normalized using (1):

$$\mathbf{X}_{ij}^{normal} = (\mathbf{X}_{ij} - \bar{\mathbf{X}}_j)/(\bar{\mathbf{S}}_j) \quad (1)$$

where $\mathbf{X}_{ij}$ denotes the $j^{th}$ input feature at the $i^{th}$ frame of the movement and $\bar{\mathbf{X}}_j$ and $\bar{\mathbf{S}}_j$ are the mean and standard-deviation of the $j^{th}$ input feature, respectively. The mean and standard deviation were calculated from the training dataset.

### *C. Deep learning models*

Two types of deep neural networks were employed: the BLSTM and transformer architectures. The models were implemented using the PyTorch deep learning framework and various additional libraries such as NumPy, Matplotlib, SciPy, Seaborn, Scikit-learn, and Optuna. Computation was run on an Nvidia Geforce RTX 4060 GPU. To reduce the number of model parameters, the markers affixed to the individual's body were divided into four segments: head (4 markers), arms (7 markers for each arm), body-pelvic (7 markers for the trunk and 4 markers for the pelvic), and legs (6 markers for each leg) (Fig 2). For each segment, separate networks were trained. To predict body posture at each frame, all four separate networks were executed simultaneously.

Each model was trained to predict the next frame based on the input data consisting of the previous 25 frames. The first 25

frames of each task needed to predict the remaining 76 frames were in-vivo measured motion capture data. During the training phase, the model employed the latest in-vivo measured data to make predictions for the next immediate observation in the time series; which is also known as the short-term prediction method. In contrast, during the evaluation phase, the predictive capability of the model was recursively utilized to forecast the entire time series. This method, known as recursive long-term prediction, used the predicted frames of the model as inputs for the next prediction step [34]. For instance, frame 26 could be predicted from frames 1 to 25, and then frame 27 could be predicted from frames 2 to 26 (considering that frame 26 was already obtained from the prediction of the model). Consequently, the prediction for each frame was dependent on the predictions made for the previous frames, enabling the generation of a complete sequence of forecasts encompassing all 101 motion frames. The architectures and methods used for training and setting the hyperparameters of the models are explained hereafter.

### *C.1. BLSTM:*

BLSTMs, bidirectional RNN variants, process sequences forward and backward (Fig. 3), outperforming unidirectional LSTMs [37], [38]. In this study, a single layer of BLSTM with 128 hidden units was used for all body segments. The final hidden state of this layer was forwarded to a linear layer that reshaped the output to the desired shape.

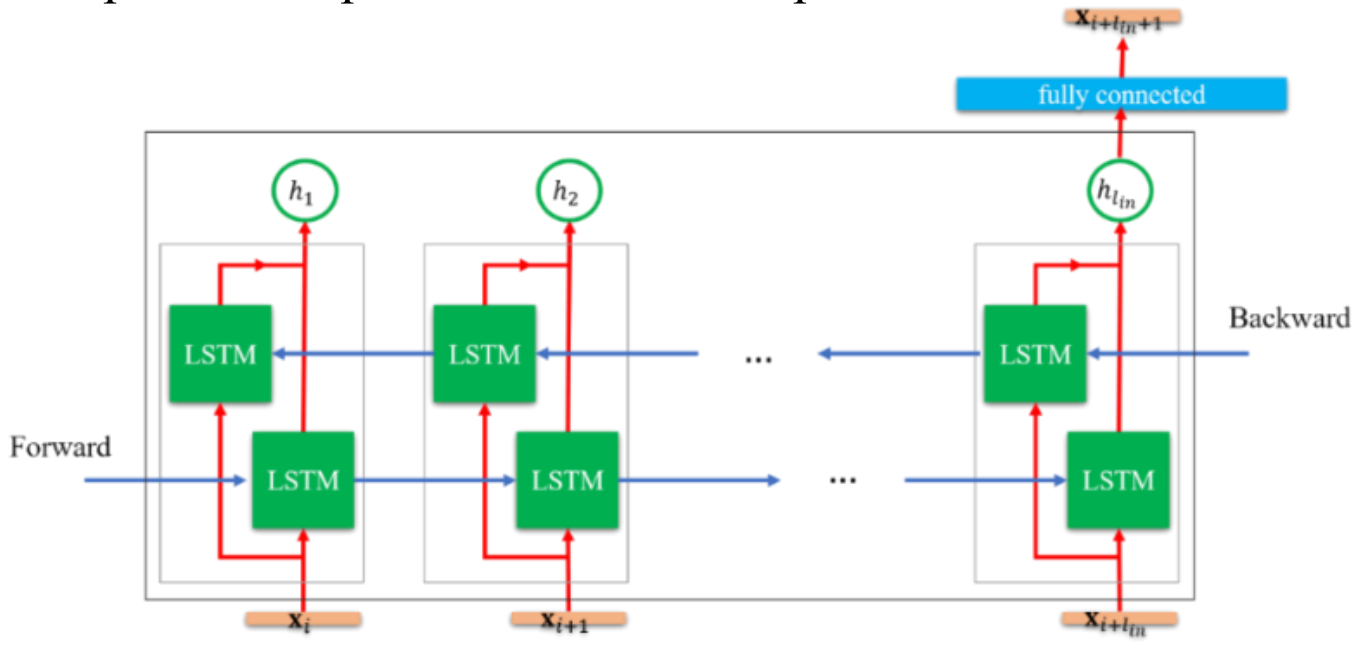

**Fig. 3.** BLSTM architecture for posture prediction. The model processes time-series data bidirectionally, capturing both past and future context [39]. Outputs from forward and backward LSTM layers are concatenated and passed to a fully connected layer to predict posture at each time step.

### *C.2. Transformer*

The transformer architecture, with encoder-decoder blocks and multi-head self-attention, has outperformed RNNs and CNNs across various fields [40], [41]. To predict the desired body kinematics, the 3D coordinates of the markers as well as seven selected features were input to the model in 25 input frames to a linear dimension expander layer. The values of this layer were then passed to the transformer encoder. Furthermore, the marker coordinates of the last frame were assigned to the linear dimension expander layer and used as input for the decoder. Finally, the decoder output was the coordinates of the markers in the next frame, following the input frames (Fig 4). The model architecture used a single encoder layer and three decoder layers, each with 16 self-attention heads. The number of feed-forward neurons were set to 512, and the decoder's output passed through a fully connected layer containing 64 neurons. A dropout probability of 0.25 was applied to the transformer. Before being fed into the transformer, the input dimensions were expanded to 96 using a linear transformation. An identical architecture was adapted for all body segments.

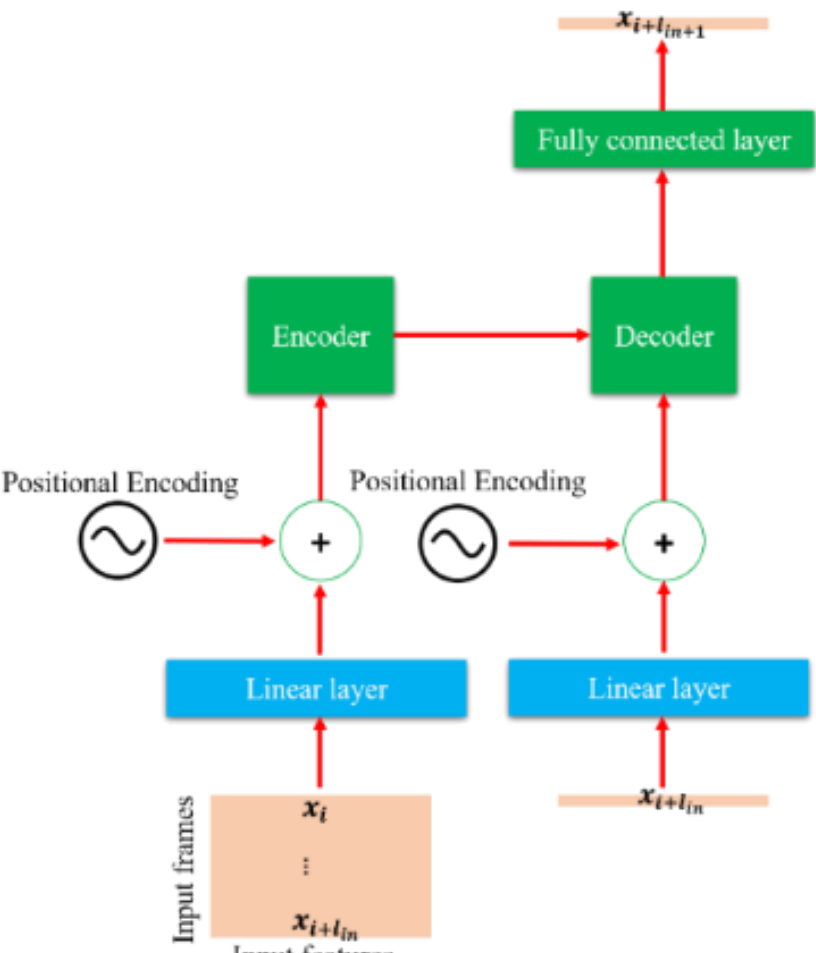

**Fig. 4.** Transformer-based model for posture prediction. Input sequences are embedded through linear layers, then processed by encoder-decoder blocks. The output is passed to a fully connected layer to generate future posture predictions using self-attention mechanisms.

### *D. Performance metrics*

The short- and long-term performances of the model were evaluated in terms of RMSE, normalized RMSE (to the range of measured data, nRMSE), and $R^2$. To calculate the RMSE, the network's output was denormalized, and the error was calculated using Eq.(2):

$$\mathrm{RMSE} = \sqrt{\frac{1}{n.f.l_{out}} \sum_{i=1}^{n} \sum_{j=1}^{f} \sum_{k=1}^{l_{out}} \left( y_{i,j,k} - \hat{y}_{i,j,k} \right)^2} \qquad (2)$$

where $n$, $f$, and $l_{out}$ represent the number of samples, number of input features, and number of output frames (1 for short-term prediction and 76 for long-term prediction), respectively. The leave-one-subject-out (LOSO) cross-validation approach was employed, ensuring that model generalization was evaluated on completely unseen individuals, which further mitigates overfitting.

### *E. Hyperparameters*

The models were trained using the Adam optimization method with parameters $\beta_1 = 0.9$, $\beta_2 = 0.98$, and $\varepsilon = 1e\text{-}9$ [42], with a mean squared error (MSE) loss function to calculate the discrepancy between one-step-ahead predictions and actual values. A leaning rate scheduler was selected by trial and error (factor = 0.1, patience = 5, minimum learning rate = 1e-7) which adjusted the learning rate based on validation loss; and early stopping was applied when no improvement was observed across multiple epochs. Model weights were updated based on this loss function and training continued for 150-200 epochs. To optimize model performance, the tree-structured Parzen estimator (TPE) algorithm [43], a Bayesian hyperparameter sampler, was employed to automatically tune hyperparameters by minimizing validation loss. The search space included dropout rate, number of attention heads, encoder/decoder layers, and other architecture parameters. Final selected values are listed in Table II. The batch sizes for the BLSTMs and transformers were set at 512. Initially, the learning rates for the BLSTMs and transformers were 0.01 and 0.001, respectively. The choices for the number of epochs, batch size, and learning rates were established based on trials and errors.

**TABLE II**. Search space for the transformer model hyperparameters

| Hyperparameter | Search space | Selected |
|---|---|---|
| **Size of expander linear layer** | [64, 96, 128] | 96 |
| **Number of self-attention heads** | [8, 16, 32] | 32 |
| **Number of feed forward neurons** | [256, 512, 1024, 2048] | 512 |
| **Encoder layers** | [1, 2, 3, 4] | 1 |
| **Decoder layers** | [1,2,3,4] | 3 |
| **Number of output neurons** | [64, 128, 256] | 64 |
| **Dropout** | [0.2, 0.25, 0.3 ,0.35, 0.4, 0.45, 0.5] | 0.25 |

### *F. Kinematic constraints*

To improve the performance of the models, and as a contribution of this study, some kinematic constraints were implemented by defining a new loss function. The length of the forearms and arms (for the arm models) and shanks (for the leg models) were considered constant during the task for each subject. To calculate the length of these segments, the Euclidean distance of the markers of each segment was calculated (Fig 2 and Table III). The new loss function ($\mathcal{L}$) was, therefore, computed using Eq.(3):

$$\mathcal{L} = \frac{1}{n.f.l_{out}} \sum_{i=1}^{n} \sum_{j=1}^{f} \sum_{k=1}^{l_{out}} \left(y_{i,j,k} - \hat{y}_{i,j,k}\right)^2 + a \times \sum_{m=1}^{M} (\ell_m - \hat{\ell}_m)^2 \tag{3}$$

The first term accounts for the loss of all markers of the model, and the second term represents the kinematic constraints for different segments in which '$m$' denotes the number of constraints of the segment model and $\ell$ is the length of the segment. Three values for coefficient $a$ were considered based on trials and errors ($a$ = 1, 10, 100).

To quantitatively evaluate the effect of this constraint, we computed the Kullback-Leibler (KL) divergence between the empirical distribution of segment lengths from the motion capture data distribution ($P$) and the predicted distributions from the models with and without the kinematic loss term, denoted $Q_{\text{kin}}$ and $Q_{\text{MSE}}$, respectively using the definition of KL divergence as Eq.(4):

$$D_{\text{KL}}(P||Q) = \sum_{l} P(l) \log \frac{P(l)}{Q(l)} \tag{4}$$

This analysis was performed over 130 movement tasks from 2 local test subjects and also 3 external test subjects.

**TABLE III**. Kinematic constraints for the segmental length in arms and legs models, see Figure. 2 b for placement of each of the markers.

| | First marker | Second marker |
|---|---|---|
| **Right upper-arm** | RHSO | RELB |
| **Left upper-arm** | LSHO | LELB |
| **Right forearm** | RELB | RWRA |
| **Left forearm** | LELB | LWRA |
| **Right shank** | RKNE | RANK |
| **Left shank** | LKNE | LANK |

## III. RESULTS

The short-term (one step ahead) and long-term (76 time-steps) predictive performance of the two trained models were evaluated. The BLSTM and transformer models were trained for 150 and 200 epochs, respectively, with early stopping of the validation MSE loss to prevent overfitting. Moreover, batch sizes of 256 and 512 for the BLSTM and transformer models, respectively, were used. The best results were obtained with $a$ = 10 for the arm model and $a$ = 1 for the leg model in the loss function (which were used for training of the final models in this study).

### *A. Performance of short-term predictions*

In the one-step-ahead prediction task, the BLSTM model consistently outperformed the Transformer model across all body segments (Table IV). The RMSE for the BLSTM model was relatively low: 0.21 mm for the body-pelvic segment, 0.31 mm for the head, 0.23 mm for the arms, and 0.25 mm for the legs. In contrast, the Transformer model exhibited higher RMSE values across all body segments, with 1.85 mm for the body-pelvic segment, 0.96 mm for the head, 0.99 mm for the arms, and 0.90 mm for the legs.

**TABLE IV.** Root mean square error (RMSE) for one-step-ahead prediction

| | RMSE (mm) | | | |
|---|---|---|---|---|
| | **Body-pelvic** | **Head** | **Arms** | **Legs** |
| **BLSTM** | 0.21 | 0.31 | 0.23 | 0.25 |
| **Transformer** | 1.85 | 0.96 | 0.99 | 0.90 |

### *B. Performance of long-term recursive predictions*

A collection of models was developed for long-term predictions by recursively utilizing one-step-ahead predictions as inputs. A total of 76 time steps in the future (until the 101st frame as the last frame) were predicted for each task. The transformer model yielded lower RMSE values (Table V), indicating its superior accuracy in comparison to the BLSTM model. This improved performance made the transformer model the preferred choice for further analysis. Its predictions are shown across multiple frames in Fig. 5, demonstrating the model's ability to capture dynamic posture changes over time with high accuracy.

**TABLE V.** Root mean square error (RMSE), normalized-RMSE (nRMSE), and $R^2$ for long-term recursive predictions

| | | RMSE (mm) | nRMSE (%) | $R^2$ |
|---|---|---|---|---|
| **BLSTM** | **Body-pelvic** | 102.3 | 16.1 | 0.97 |
| | **Head** | 117.9 | 11.9 | 0.98 |
| | **Arms** | 114.7 | 9.3 | 0.96 |
| | **Legs** | 67.9 | 18.2 | 0.93 |
| | **Whole-body** | 100.0 | 14.5 | |
| **Transformer** | **Body-pelvic** | 36.7 | 5.7 | 0.99 |
| | **Head** | 53.5 | 4.5 | 0.99 |
| | **Arms** | 49.4 | 4.1 | 0.99 |
| | **Legs** | 28.7 | 7.5 | 0.99 |
| | **Whole-body** | 41.4 | 5.7 | |
| **Transformer (with kinematic constraints)** | **Arms** | 45.4 | 3.9 | 0.99 |
| | **Legs** | 22.7 | 5.8 | 0.99 |
| | **Whole-body** | 37.2 | 5.0 | |

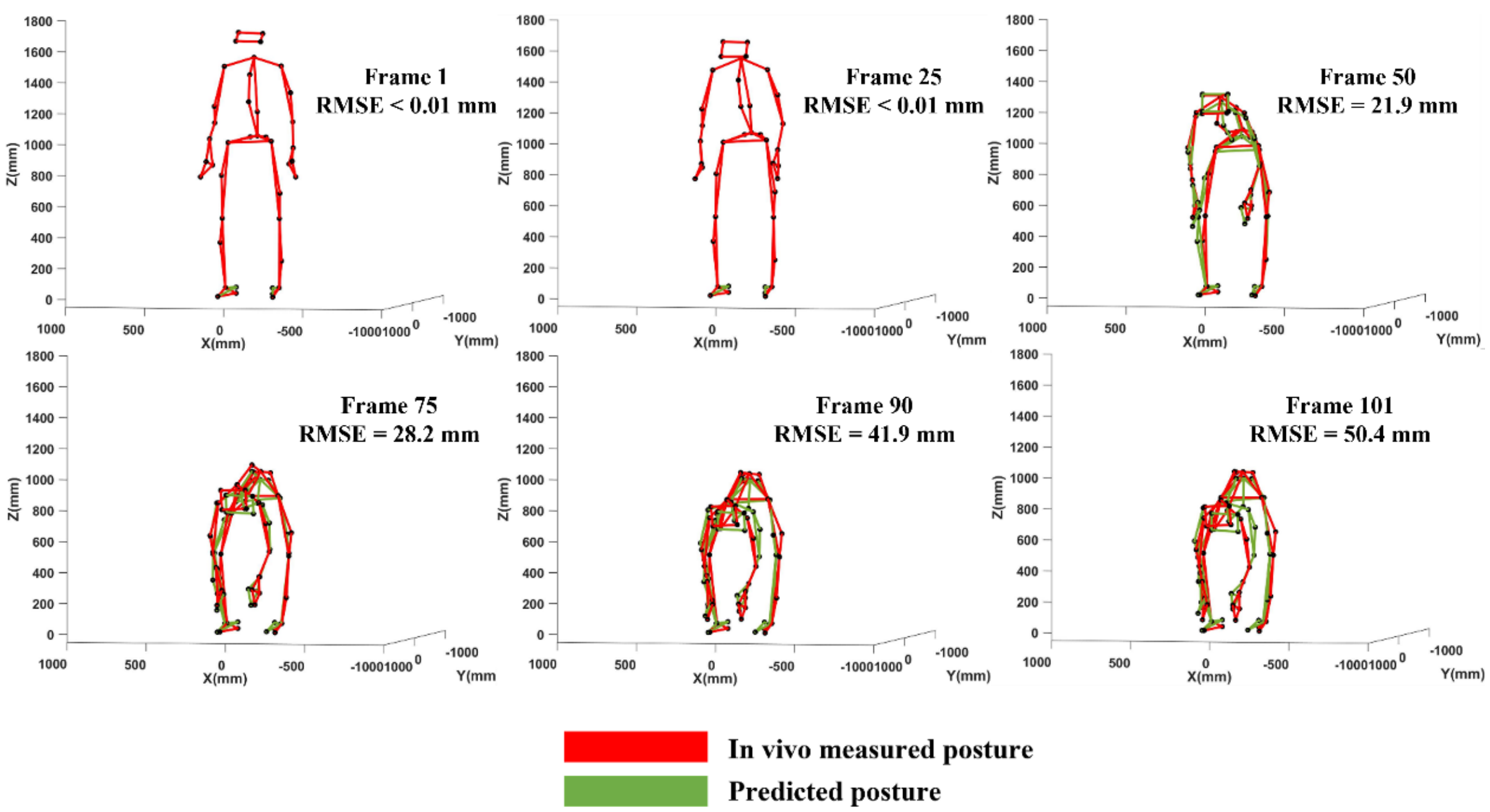

**Fig. 5.** Predicted vs. actual dynamic postures across six key time frames using the Transformer model. Each subplot shows a comparison between the predicted posture (green) and the ground truth (red) during a load-reaching task at different time frames (1, 25, 50, 75, 90, and 101). The associated RMSE values (in mm) indicate the prediction accuracy at each frame. The task involved stoop lifting with both hands, targeting a final load position at coordinates (x = 0 mm, y = 580.6 mm, z = 128.3 mm). The subject's body dimensions were 70 kg and 180 cm.

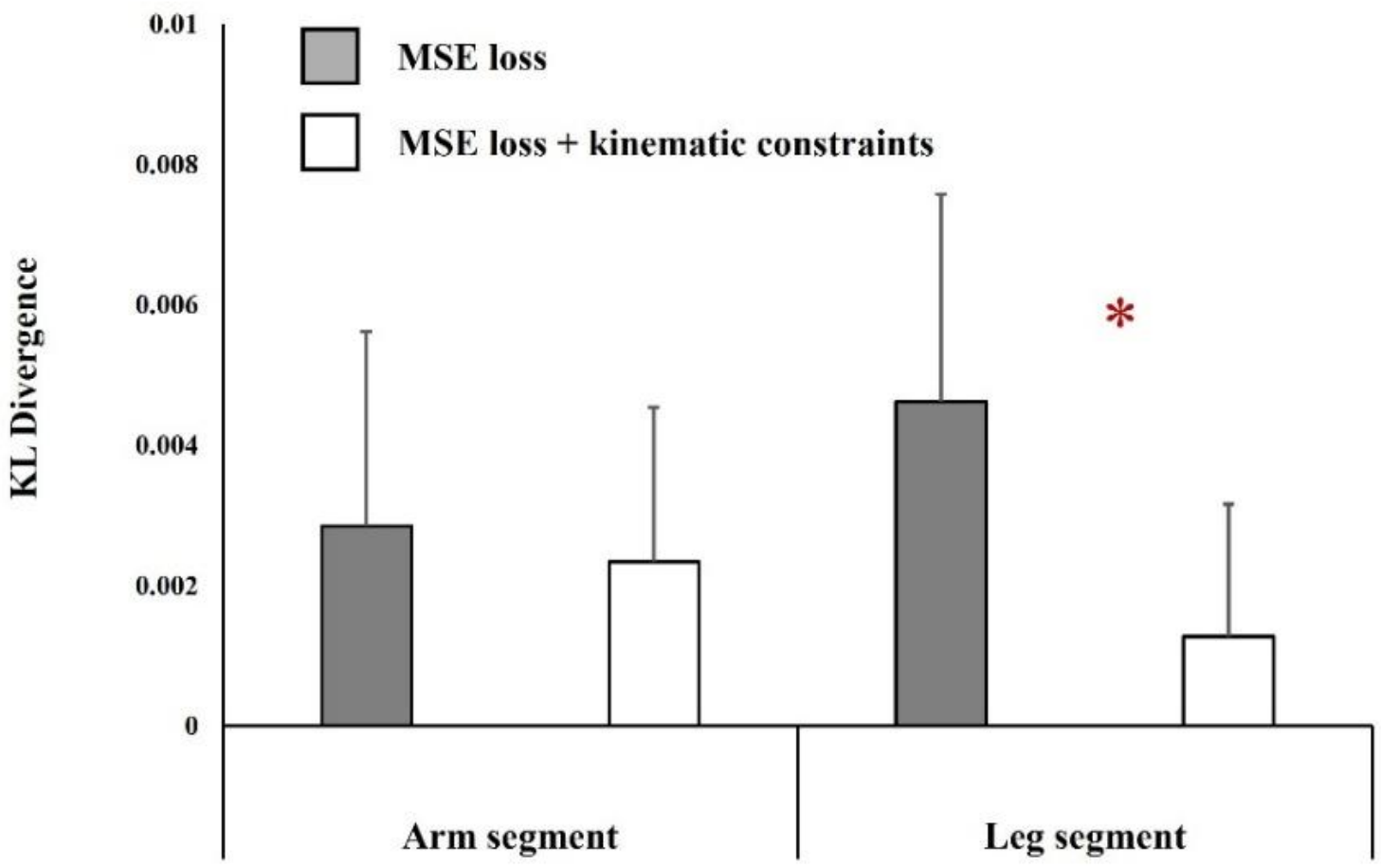

**Fig. 6.** KL divergence values for arm and leg segments under two training conditions: models trained with MSE loss only (gray bars) and models trained with MSE loss plus kinematic constraints (white bars). Error bars indicate standard deviation. A significant reduction in KL divergence for the leg segment when using kinematic constraints is denoted by the red asterisk ($p = 0.0226$). This analysis was performed over 130 movement tasks from 2 local test subjects and 3 external test subjects. The results show that the model with the kinematic loss exhibited significantly lower KL divergence compared to the MSE-only model, indicating a closer match to the real segment length distributions.

### *C. Effect of kinematic constraints on the performance of the networks*

The inclusion of kinematic constraints improved the performance of the transformer-based model, i.e., the RMSEs decreased by approximately 4 and 6 mm for the neural networks of the arms and legs, respectively (Table V). The results (Fig. 6) showed that the model with the kinematic loss $Q_{\text{kin}}$ exhibited significantly lower KL divergence compared to the MSE-only model, indicating a closer match to the real segment length distributions. Furthermore, prediction accuracy (as measured by standard joint position error metrics) was also improved. This demonstrates that kinematic constraints not only yield more biomechanically plausible outputs but also enhance overall prediction performance. This improvement underscores the importance of maintaining realistic anatomical constraints during dynamic posture prediction.

### *D. Leave-one-subject-out cross-validation (LOSO)*

Based on the findings presented in section III. B, the transformer model was selected as the final model. To evaluate the network's generalizability, the LOSO method was implemented, and its performance was presented in Table VI. Variations in RMSE values were observed across the body segments, with the highest RMSE recorded for the body-pelvic segment (60.4 mm) and the lowest for the legs (42.3 mm). Despite the variability across subjects, the overall performance of the Transformer model is shown to demonstrate its robustness and ability to generalize to different individuals.

### *E. External test dataset*

To further examine the generalizability of the developed models, an additional evaluation was performed using the data from three external subjects not included in the original dataset. These participants had an average weight of 71.9 ± 8.4 kg and height of 178.3 ± 5.9 cm. Similar load-reaching tasks were conducted using the same marker setup and protocol described in Section II.A. The transformer model with kinematic constraints was applied without retraining or fine-tuning. The performance metrics (RMSE, nRMSE, and $R^2$) for the arms and legs were calculated, as summarized in Table VII.

## IV. Discussion

This study aimed to develop deep learning models for the dynamic prediction of the human whole-body posture during load-reaching tasks. For this purpose, the full-body plug-in gait markers were divided into four segments. Subsequently, short-term prediction (one-step-ahead) models were trained for each segment based on the BLSTM and transformer architectures. The inputs to the models were the first 25 frames of motion as well as seven simple task- and subject-related characteristics. All motion frames during a load-reaching task could be predicted recursively using the previous frames. The results indicated that based on our tuned hyperparameters, in short-term predictions, the BLSTM model outperformed the transformer (Table IV). However, for long-term predictions, the transformer model was more accurate than the BLSTM (Table V). Despite the relatively high short-term accuracy of the BLSTM, we decided to use our transformer model because of its relatively high accuracy in the long term.

### *A. Analysis of results*

The increase in the error of long-term predictions could be related to the training process. During the training phase of the recurrent neural networks, all model inputs were true values; however, during the testing of these networks, the prediction of the network, which has some error itself, was input to the model. The accumulation of these errors during the next step of the predictions caused the network to have more errors as the number of predicted frames increased. Therefore, the predicted value further deviated from the actual outputs in the previous frames. This issue is known as the "teacher forcing" or "exposure bias" in recurrent networks [44], [45]. One solution to this issue is to use a technique called "scheduled sampling" [46]. This involves training the model using both the actual input data and the predictions made by the model, allowing for more accurate training and thus improved results. Another solution to improve the performance of the BLSTM network is use of a cost function to optimize or evaluate the long-term in addition to optimize one-step ahead prediction. Since only the short-term prediction was minimized in the cost function of the network, it could be expected that the network was over-fitted on the short-term prediction. Previous studies [29] used dynamic time warping to choose the model with superior long-term prediction capabilities while simultaneously optimizing the MSE or RMSE errors. It was found that while LSTM models outperformed transformer models in short and long-term predictions, they were the most affected by Gaussian noise [29]. These results suggest that transformers are more resilient to noises than LSTMs and can be used for predicting body posture.

**TABLE VI**. Root mean square error (RMSE) for the leave-one-subject-out (LOSO) cross validation

| | **RMSE: mean ± std (mm)** | | | | |
|---|---|---|---|---|---|
| **Transformer** | **Arms** | **Legs** | **Head** | **Body-pelvic** | **Whole-body** |
| | 52.3 ± 12.6 | 42.3 ± 28.7 | 53.6 ± 18.2 | 60.4 ± 29.4 | 54.4 ± 17.6 |

**Table VII**. Performance of the Transformer model on biomechanical prediction for 3 external test subjects, reported using RMSE, nRMSE, and R². Results are shown for both the baseline model and the version with kinematic constraints, which improved arm segment prediction accuracy.

| | | **RMSE (mm)** | **nRMSE (%)** | **$R^2$** |
|---|---|---|---|---|
| **Transformer** | **Body-pelvic** | 106.2 | 13.1 | 0.97 |
| | **Head** | 126.8 | 12.5 | 0.97 |
| | **Arms** | 98.8 | 7.8 | 0.96 |
| | **Legs** | 64.1 | 19.2 | 0.93 |
| | **Whole-body** | 95.5 | 13.8 | 0.97 |
| **Transformer (with kinematic constraints)** | **Arms** | 84.8 | 6.8 | 0.97 |
| | **Legs** | 63.5 | 20.5 | 0.93 |
| | **Whole-body** | 88.4 | 13.0 | 0.97 |

It is challenging to directly compare findings of the present study to those reported in the literature since no prior studies have investigated the use of BLSTM and transformer methods for posture prediction during MMH activities. Furthermore, previous studies have investigated different tasks, such as gait cycles [27], [28], [29], [47] or have used multi-layer perceptron (MLP) models to predict dynamic postures during MMH activities [12], [17], [18], [20], [21]. In agreement with one previous study [29], our findings indicated that BLSTM outperformed the transformer in short-term predictions. In contrast, we found that our transformer network had superior performance in long-term predictions as compared to the BLSTM model. Furthermore, in our previous model [18], posture prediction was performed using a frame-wise MLP approach based on static task- and subject-related inputs together with the frame index, resulting in a 41.5 ± 5 mm whole -body RMSE under LOSO cross-validation. In contrast, the current transformer model formulates posture prediction as a time-series forecasting problem, where the remaining frames are predicted recursively after the first 25 measured frames. Because predicted frames are recursively used as inputs for subsequent steps, errors may accumulate over time, leading to a higher overall RMSE (54.4 ± 17.6 mm). However, unlike the frame-wise MLP model, the transformer captures temporal dependencies across frames, while biomechanical consistency is enforced through the defined loss function rather than the architecture itself, helping maintain temporally coherent motion despite the increased forecasting difficulty.

The LOSO cross-validation results showed higher variability for the leg segment (42.3 ± 28.7 mm), which reflects substantial inter-subject differences in lower-limb kinematics during load-reaching tasks. Because LOSO evaluates strict subject-independent generalization, the model is required to predict motion for unseen individuals with different anthropometric characteristics and movement strategies. Variations in stance configuration, knee flexion patterns, and compensatory adjustments contribute to increased prediction uncertainty in the lower extremities. Although the relative standard deviation is large, the absolute error magnitude remains within a physiological range for dynamic whole-body posture prediction[18], [19], [20] . These findings suggest sensitivity to inter-subject variability rather than model failure and highlight the importance of future subject-adaptive or larger-scale training strategies to further improve robustness.

To contextualize the magnitude of the reported posture prediction errors, the overall RMSE in this study was 37.2 mm for the internal train–validation–test subjects, 54.4 ± 17.6 mm under LOSO cross-validation, and 88.4 mm for the external subjects. A prior modeling study [19] investigating the effect of posture-prediction errors on musculoskeletal model outputs reported that an overall posture prediction error of 74 mm led to spinal load estimations that closely agreed with those derived from measured postures. These findings suggest that posture prediction errors within this range can still produce acceptable biomechanical outputs when used as inputs to musculoskeletal models. Therefore, the prediction errors obtained in the present study for the internal subjects (37.2–54.4 mm) remain within a range that can reasonably support downstream ergonomic and musculoskeletal analyses, although the larger error observed for the external subjects highlights the need for larger and more diverse datasets to further improve model generalization.

The present study also proposed a new method for improving the accuracy of posture prediction models. The approach involved adding terms to the traditional MSE loss functions to maintain the length of body segments at their initial measured values. By incorporating these terms into the loss function, RMSEs of the arm and leg models improved from 49 to 45 mm (8% improvement) and from 28 to 22 mm (21% improvement). While this new loss function improved the accuracy of the models, its performance for other prediction models remains to be investigated.

### B. Limitations

There were some limitations to the current study. First, generating the first 25% of an individual's movement during load-reaching activities in this study relied on measured motion capture data to evaluate the intrinsic forecasting capability of the proposed temporal models. However, in real-world applications where such measured anatomical posture data may not be available at initialization, these initial frames would need to be estimated. In that context, our previously developed whole-body posture prediction model [18], which predicts posture from static task- and subject-related inputs, could be used to generate the first segment of motion before applying the proposed time-series models. Second, the collected dataset only included young and novice males, i.e., the generalizability of the models might not be valid for older, female, or experienced people. Additionally, the data were only applied to normal-weight, right-handed subjects, which means that the predictions may not be accurate for people with a body mass index (BMI) less than 18 or over 26 or left-handed individuals. Furthermore, the study exclusively focused on load-reaching tasks and did not address load-moving or load-leaving tasks. The tasks that could be performed with only one hand were limited to the use of a stoop lifting technique, and did not include the full squat and semi-squat lifting techniques. Moreover, the full squat and semi-squat lifting techniques were only tested with loads up to a height of 30 cm, making it challenging to anticipate scenarios involving higher load heights.

## V. Conclusion

This study is significant because of its use of deep learning to capture time series dependencies in 3D motion, offering a distinctive approach to understanding motion dynamics. This study focused on investigating the use of deep neural networks for predicting human postures during load-reaching activities using two time-series models: the BLSTM and transformer architectures. The models used various input parameters including load position, adapted lifting and handling techniques, body anthropometry, and early task period data to predict body coordinates for the later task period. Moreover, a new method was introduced to enhance the accuracy of posture prediction models by optimizing a novel loss function to maintain constant body segment lengths. While in short-term predictions, our BLSTM model outperformed the transformer model, for long-term predictions, our transformer model was more accurate than the BLSTM model.

## Appendix A

### *Abbreviations*

Bidirectional long short-term memory (BLSTM), Body mass index (BMI), Computer vision (CV), Convolutional neural networks (CNNs), Fully connected neural networks (FCNs), Handling technique (HT), Leave-one-subject-out (LOSO), Lifting technique (LT), Long short-term memory (LSTM), Machine learning (ML), Manual material handling (MMH), Mean squared error (MSE), Multi-layer perceptron (MLP), One-dimensional (1D), Recurrent neural networks (RNNs), Root mean square error (RMSE), Three-dimensional (3D)

## Acknowledgment

We appreciate the assistance of members of the Djawad Movafaghian Research Center in Rehab Technologies and Social Robotics Lab at Sharif University of Technology.

## Code and Data Availability

We also provide the full implementation of the proposed models and training pipeline for public access at: [https://github.com/Niloufar-Husseini/posture-prediction-models].